\documentclass[journal]{IEEEtran}

\usepackage{hyperref}
\usepackage{graphicx}
\usepackage{subfigure}
\usepackage[flushleft]{threeparttable}
\usepackage{booktabs} 
\usepackage{amsfonts}
\usepackage{multirow}
\usepackage{tabularx}
\usepackage{xcolor}
\usepackage{footnote}
\usepackage{caption}
\usepackage{textcomp, gensymb}
\usepackage{amsmath}

\newcolumntype{M}{>{$}c<{$}}
\newcolumntype{Z}{>{\centering\arraybackslash}X}
\newcolumntype{C}{>{\centering\arraybackslash}p}
\newcolumntype{Y}{>{\centering\arraybackslash}X}
\newcolumntype{L}{>{\raggedright\arraybackslash}p}

\newcommand{\specialcell}[2][c]{%
  \begin{tabular}[#1]{@{}c@{}}#2\end{tabular}}

\ifCLASSINFOpdf
\else
\fi

\begin{document}
\title{Child PalmID: Contactless Palmprint Recognition}

\author{Anil~K.~Jain,~\IEEEmembership{Life~Fellow,~IEEE},~Akash~Godbole,~Anjoo~Bhatnagar~and~Prem~Sewak~Sudhish
}


\maketitle

\begin{abstract}

Developing and least developed countries face the dire challenge of ensuring that each child in their country receives required doses of vaccination, adequate nutrition and proper medication. International agencies such as UNICEF, WHO and WFP, among other organizations, strive to find innovative solutions to determine which child has received the benefits and which have not. Biometric recognition systems have been sought out to help solve this problem. To that end, this report establishes a baseline accuracy of a commercial contactless palmprint recognition system that may be deployed for recognizing children in the age group of one to five years old. On a database of contactless palmprint images of one thousand unique palms from 500 children, we establish SOTA authentication accuracy of 90.85\% @ FAR of 0.01\%, rank-1 identification accuracy of 99.0\% (closed set), and FPIR=0.01 @ FNIR=0.3 for open-set identification using PalmMobile SDK from Armatura.

\end{abstract}


\IEEEpeerreviewmaketitle

\section{Introduction}

\IEEEPARstart{I}{n} 2020, 22\% of the world's children were physically stunted due to malnourishment and lack of adequate medication\footnote{https://www.who.int/data/gho/data/themes/topics/joint-child-malnutrition-estimates-unicef-who-wb}. A majority of these children live in developing countries where healthcare facilities and other resources are not readily available. Furthermore, these countries do not have any secure government-issued identification documents to verify the recipient of the services and curtail the occurrence of fraud. To solve this problem, many international organizations such as the World Health Organization, Bill and Melinda Gates Foundation and the World Food Programme have made substantial efforts to reduce the rate of malnourishment as well as improve vaccination coverage among the more vulnerable population of the least developed countries.
\par One initiative that has received significant attention is the use of biometrics (e.g., fingerprints and palmprints) for large-scale and accurate identification of children.
However, biometrics-based identification solutions for children have yet to meet the field operation deployment criteria in many ways, including low cost biometric acquisition, high longitudinal recognition accuracy and robustness to imaging environment. Indeed, large-scale biometric identification systems in use today fail to account for children, arguably the most vulnerable population. The largest civil biometric database in the world,
Aadhaar, only enrolls subjects over the age of 5 \cite{uidai}. This leaves a population of almost 118 million children unaccounted for in India alone . Further, India alone accounts for 25 million newborns every year.
\par It is important to keep in mind that a biometric trait must meet the \textit{persistence} and \textit{individuality} requirements for the population under consideration \cite{jainbiometrics2011}. In our application, this means that the recognition accuracy of the biometric trait should not change over time and the biometric trait is different for different children. This rules out face biometrics since the face of a child dramatically changes during the first few years of their life. Footprints (friction ridge pattern at the bottom of the feet) do not satisfy the real-time acquisition requirement; the child's socks and footwear would need to be removed, or in cases where the child is barefooted, it may need to be cleaned to capture good quality footprints. Iris images are difficult to capture if the child is sleeping or crying. Further, capturing iris is akin to an ophthalmic exam which may make the parents uneasy. 
These limitations, paired with the rise of infectious diseases in the world, has motivated a push to develop biomteric systems that do not require physical contact with any surface\footnote{https://one.amazon.com/}. Toward this end, while contactless palmprint recognition has been studied primarily for adults, we propose it as a cost effective solution for large-scale child identification. Indeed, there is no requirement of custom sensors since smartphone cameras have sufficient resolution to capture contactless palmprint images.

\begin{figure}[t]
\begin{center}
\includegraphics[width=\linewidth]{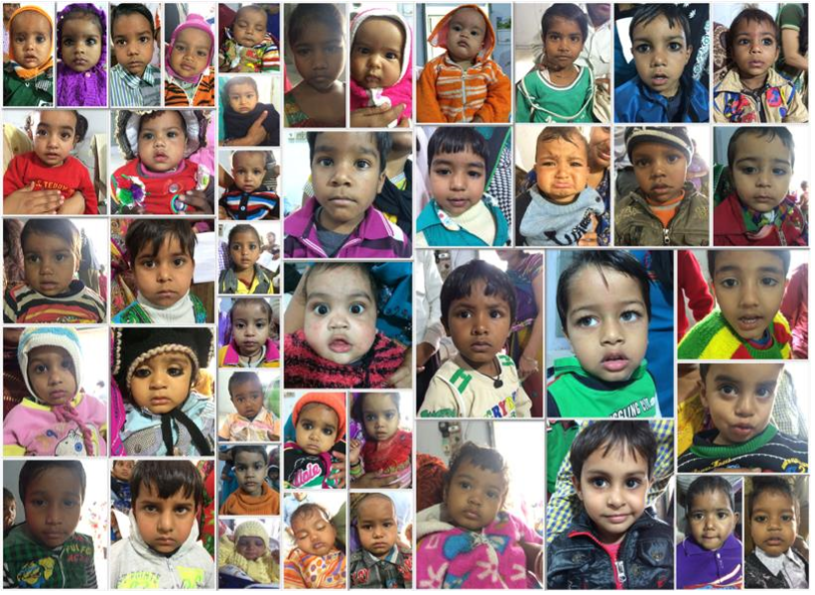}
\caption{Faces of infants, toddlers and preschoolers (0-5 years of age) in our database.}
\label{fig:toddlers}
\end{center}
\end{figure}

\begin{figure}[t]
\begin{center}
\includegraphics[width=\linewidth]{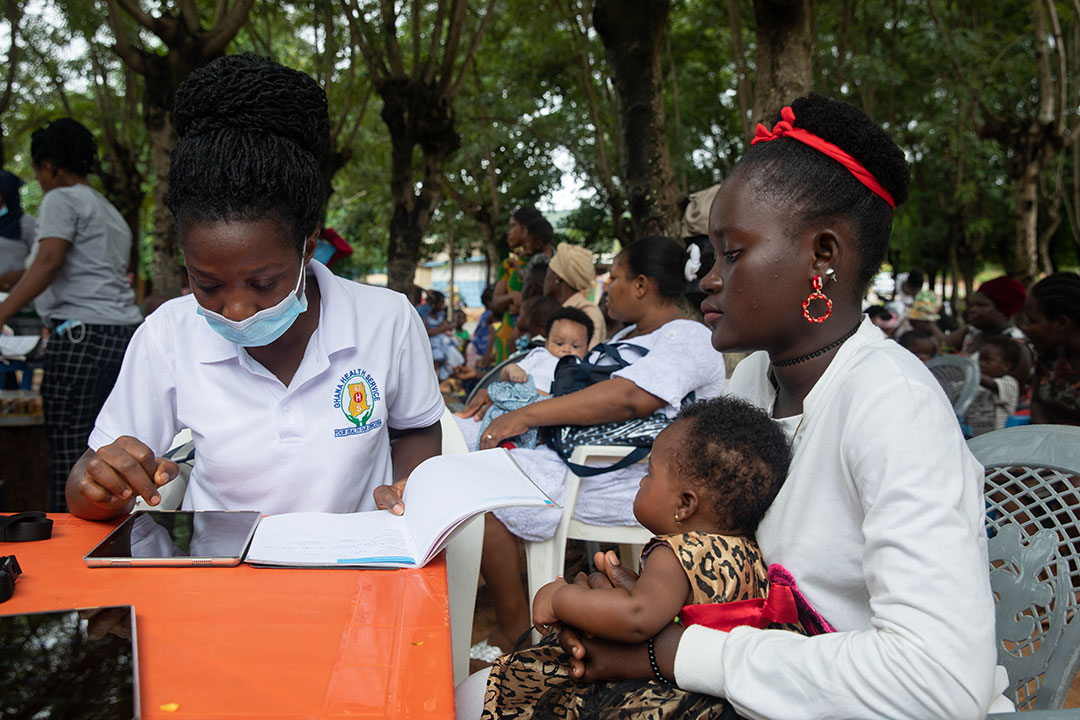}
\caption{Vaccination clinic organized by Gavi, The Vaccine Alliance, in Africa \cite{gavi}}
\label{fig:vaccine_drives}
\end{center}
\end{figure}

\begin{figure}[t]
\begin{center}
\includegraphics[width=\linewidth]{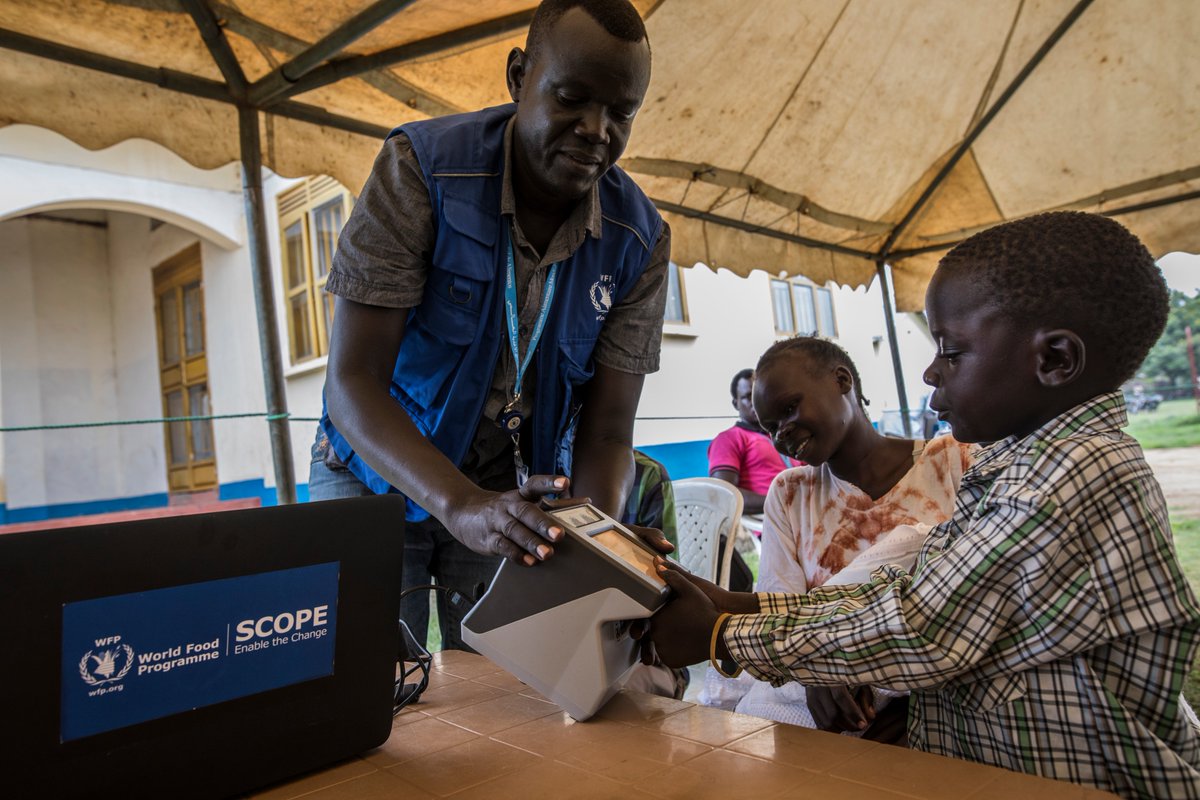}
\caption{WFP biometric enrollment of children at food distribution center in South Sudan \cite{programme_2018}.}
\label{fig:wfp}
\end{center}
\end{figure}

\par Of the prominent biometric traits, we believe that friction ridge patterns (fingerprint, palmprint and footprint) are the most promising for child recognition. Friction ridge patterns are considered to be (i) unique, (ii) present at birth, and (iii) stable over time in terms of recognition accuracy.  
Footprint recognition is inconvenient due to the reasons aforementioned and fingerprint recognition has not yet been able to deliver the required recognition accuracy levels \cite{engelsma2021infant}. Palmprints, on the other hand, have the advantages of a larger surface area compared to fingerprints. 

\par Table \ref{table:previous_work} shows some of the studies conducted in the field of biometric recognition for children using different modalities. The fingerprint modality has been the popular choice thus far but recent studies have shown a trend towards contactless palmprints. The primary obstacle in contactless palmprint recognition for children is that of data.


\par Prior attempts at palmprint-based recognition for children focused on newborns and infants (less than 12-months old). These attempts were largely unsuccessful because of the challenges in capturing palmprints of such young children who are not “cooperative” in terms of understanding instructions for placing their hand over palmprint readers \cite{lemes2011biometric}. To keep the child recognition problem tractable, we focus on children between 1 to 5 years. Child development studies report that starting at the age of one, a child can follow instructions and be considered as a “cooperative” subject in terms of opening the fist and placing the palm over the palmprint reader (in our case a mobile phone camera). This age group is also of interest to Aadhaar 2.0 \cite{press_info}, where one of the objectives is to lower the enrolment age which has been set at 5 since the inception of the program in 2009.

\section{Contactless Palmprint Recognition}

\begin{table*}[]
\caption{Summary of biometrics-based literature on child (infants and toddlers) identification}
\centering 
\begin{threeparttable}
\begin{tabular}{|C{0.1\linewidth}|C{0.25\linewidth}|C{0.1\linewidth}|C{0.2\linewidth}|C{0.2\linewidth}|}
\noalign{\hrule height 1.0pt}
\specialcell{\textbf{Authors}}  & \specialcell{\textbf{Database collection medium}} & \specialcell{\textbf{Modality}} & \specialcell{\textbf{Age Group (\# of subjects)}} & \specialcell{\textbf{Conclusion}} \\
\noalign{\hrule height 1.0pt}
Jain et al., 2016 \cite{jain2016fingerprint}   & Contact-based commercial and custom fingerprint sensors  & Fingerprint & 0-5 years (309) & High accuracy shown by commercial matchers \\
\noalign{\hrule height 0.5pt}
Liu, 2017 \cite{liu2017infant}  & Contact-based commercial fingerprint sensor & Footprint & 1-9 months (60) & Footprint recognition is feasible as a biometric for children\\
\noalign{\hrule height 0.5pt}
Ramachandra et al., 2018 \cite{ramachandra2018verifying} & Smartphone & Palm & 6-36 hours (50) & Strong recognition results using transfer learning \\
\noalign{\hrule height 0.5pt}
Yambay et al., 2018 \cite{yambay2019feasibility} & Commercial contact-based footprint sensor & Toe print & 4-13 years (177) & Strong recognition performance using commercial matchers \\
\noalign{\hrule height 0.5pt}
Saggese et al., 2019 \cite{saggese2019biometric} & Custom contactless fingerprint sensor & Fingerprint & 0-18 months (504) & Competitive performance using hand-crafted features \\
\noalign{\hrule height 0.5pt}
Engelsma et al., 2021 \cite{engelsma2021infant}  & Custom contactless fingerprint sensor & Fingerprint & 8-16 weeks (315) & High recognition accuracy using fingerprints \\
\noalign{\hrule height 0.5pt}


\end{tabular}
\end{threeparttable}
\label{table:previous_work}
\end{table*}
\begin{figure}[t]
\begin{center}
\includegraphics[width=\linewidth]{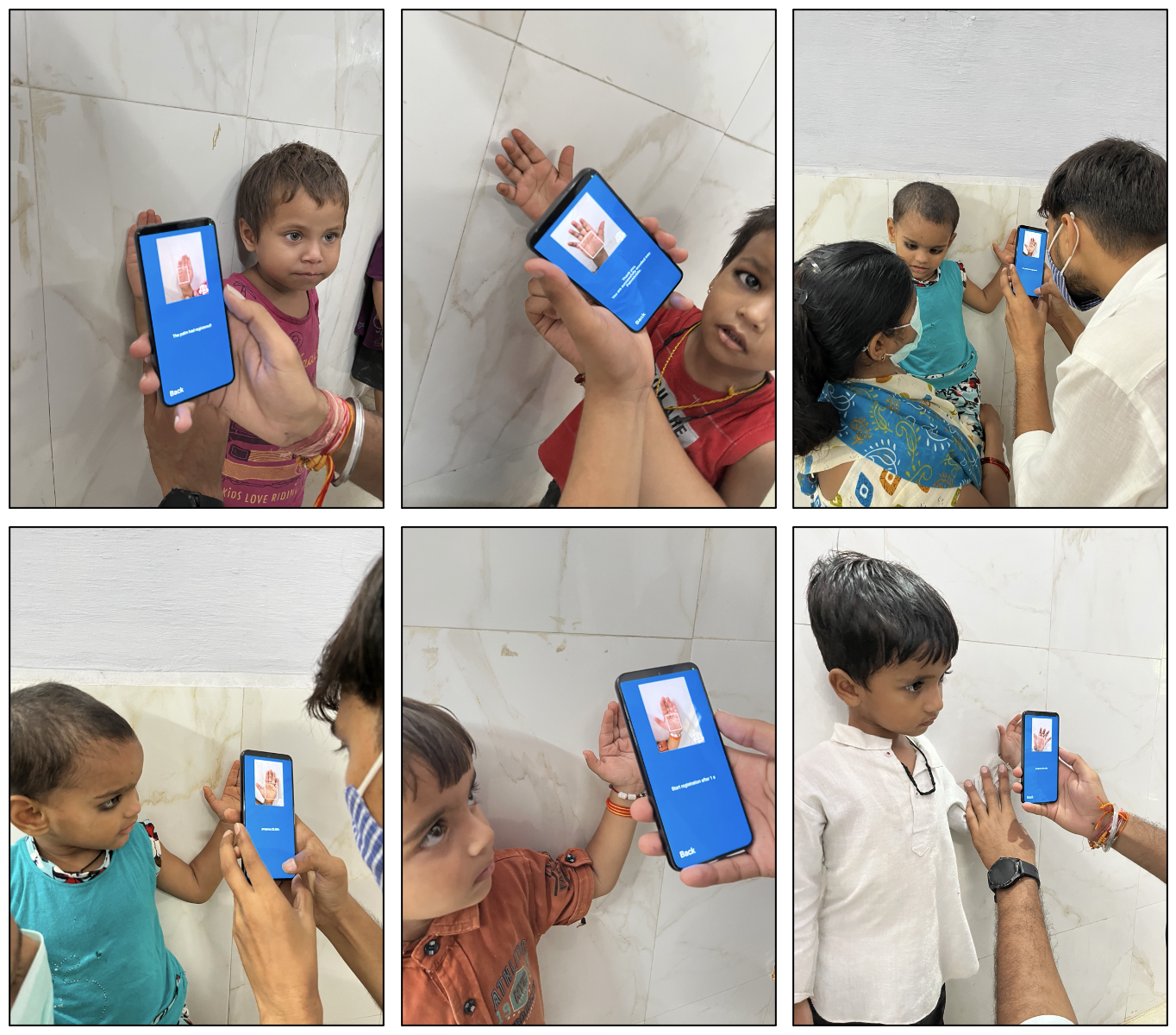}
\caption{Contactless palmprint capture process of five different subjects  using PalmMobile SDK at Saran Ashram Hospital, Dayalbagh, Agra, India. The trained volunteer kept an open communication with the child while collecting the data to make them feel comfortable. In some cases, the parent of the child also assisted in keeping the hand of the child steady. The orientation of the phone had to be adjusted based on the placement of the palm that the child is comfortable with.}
\label{fig:data_collection}
\end{center}
\end{figure}
\par Palmprint recognition broadly consists of two stages: (i) finding a "region of interest (ROI)", the central part of the palm and (ii) finding the similarity between two ROIs extracted from two different palm images. Current deployments of contactless palmprint recognition systems \cite{arcieri_2021, publisher_2022, matrix_access_control_2022} are predominantly geared towards adults \cite{dian2016contactless, zhang2017towards, liu2020contactless, morales2011towards, wu2014sift, leng2017dual, mascellino_2021}. In constrast, there are an extremely limited number of studies done for young children under the age of 5 \cite{rajaram2022palmnet}. There are no contactless palmprint datasets for children available in the public domain, nor any other known sources. Additionally, due to stringent privacy laws surrounding children, it is difficult to obtain such a database. This makes it extremely challenging to train a deep-network based system that relies on large amounts of data to achieve a competitive accuracy.

\par 


\subsection{ROI Extraction:}
\par ROI extraction is a critical part of palmprint recognition. There have been many studies focused on solely extracting a robust region of interest from a palmprint image. These methods range from using handcrafted features to using deep networks. 
\par The main drawback of handcrafted features is the constraints within which the input image must be captured. For instance, in \cite{zhang2017towards}, the palmprint is first binarized and the contours corresponding to the finger valleys are extracted. Using these points, geometrical operations are performed that yield a consistent ROI. The drawback of this approach is that it is assumed that the fingers will have a great deal of separation between them to facilitate a robust binary mask. So, if the user does not stretch the fingers of their hand wide apart, the ROI extracted will not be of a good quality.
\par Another approach is using a state of the art CNN architecture (e.g. ResNet, VGG, Inception, etc.) that can learn to extract a robust region of interest. It is typically carried out by annotating a training dataset for landmark points and training a network that can generalize to as many orientations of the palm as possible. The main benefit of this method is that the fast extraction of features and matching compared to handcrafted features. However, handcrafted features may generalize better than deep networks on other datasets.

\subsection{ROI matching:}
\par Traditional methods of matching extracted ROIs rely on handcrafted image processing features such as SIFT descriptors and CompCode maps using Gabor filters.
\par Once the ROI is extracted, the matching system is developed by fine-tuning a pre-existing deep network on the extracted ROIs \cite{zhang2017towards, liu2020contactless}. Once trained, the embeddings from these networks are used as feature sets for matching - providing speed and security.
This task is dependent on the robustness of the ROI extraction module. Using a deep network embedding, obtaining a similarity score becomes as simple as computing a scalar product between two features which facilitates a large throughput of the system.

\begin{figure*}[t]
\begin{center}
\includegraphics[scale=0.4]{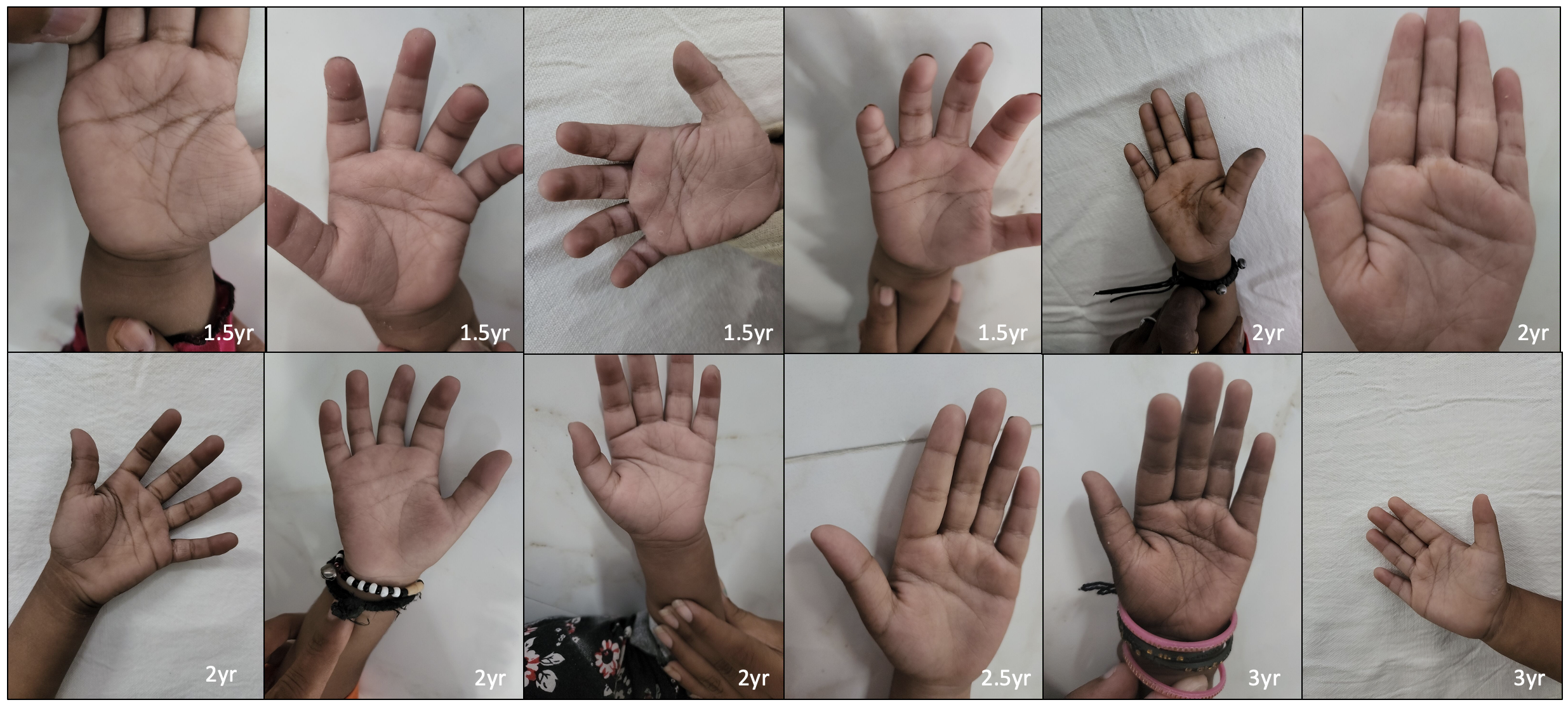}
\caption{Sample images in Child PalmDB. The image quality and palm orientation vary depending on the level of cooperation of the child. In some cases, either the operator or the parent assisted the child in opening their hand so that the palmprint is clearly visible.}
\label{fig:collage}
\end{center}
\end{figure*}

\begin{figure}[t]
\begin{center}
\includegraphics[scale=0.45]{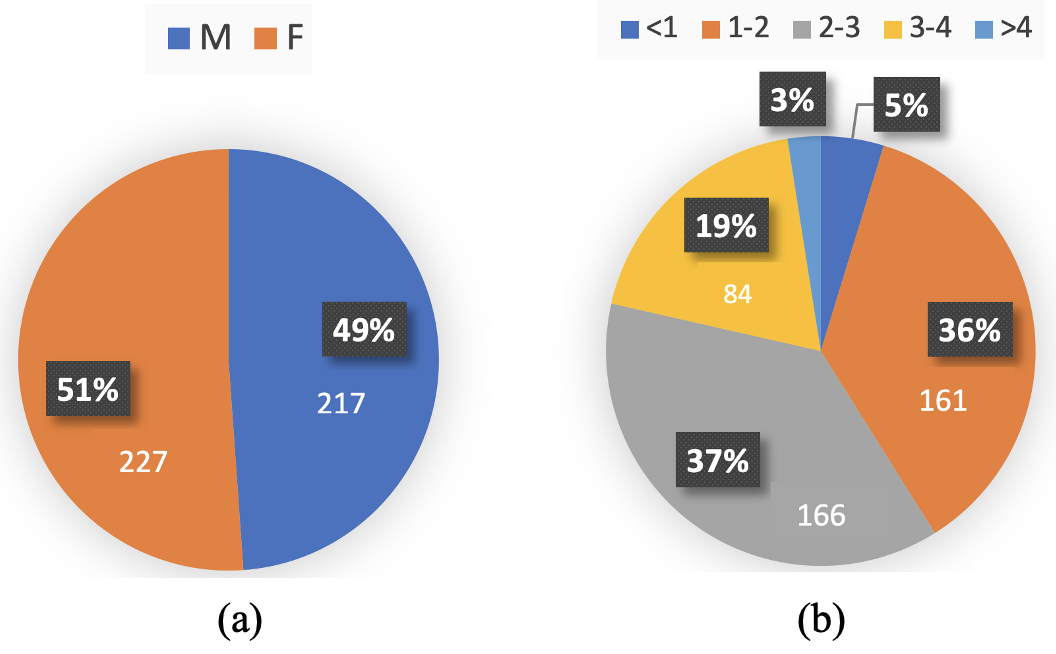}
\caption{Distribution of (a) gender and (b) age of subjects in the contactless palm image dataset, Child PalmDB. These attributes are available for 444 subjects out of a total of 515 subjects.}
\label{fig:demographics}
\end{center}
\end{figure}

\begin{figure}[t]
\begin{center}
\includegraphics[scale=0.3,angle=270,origin=c]{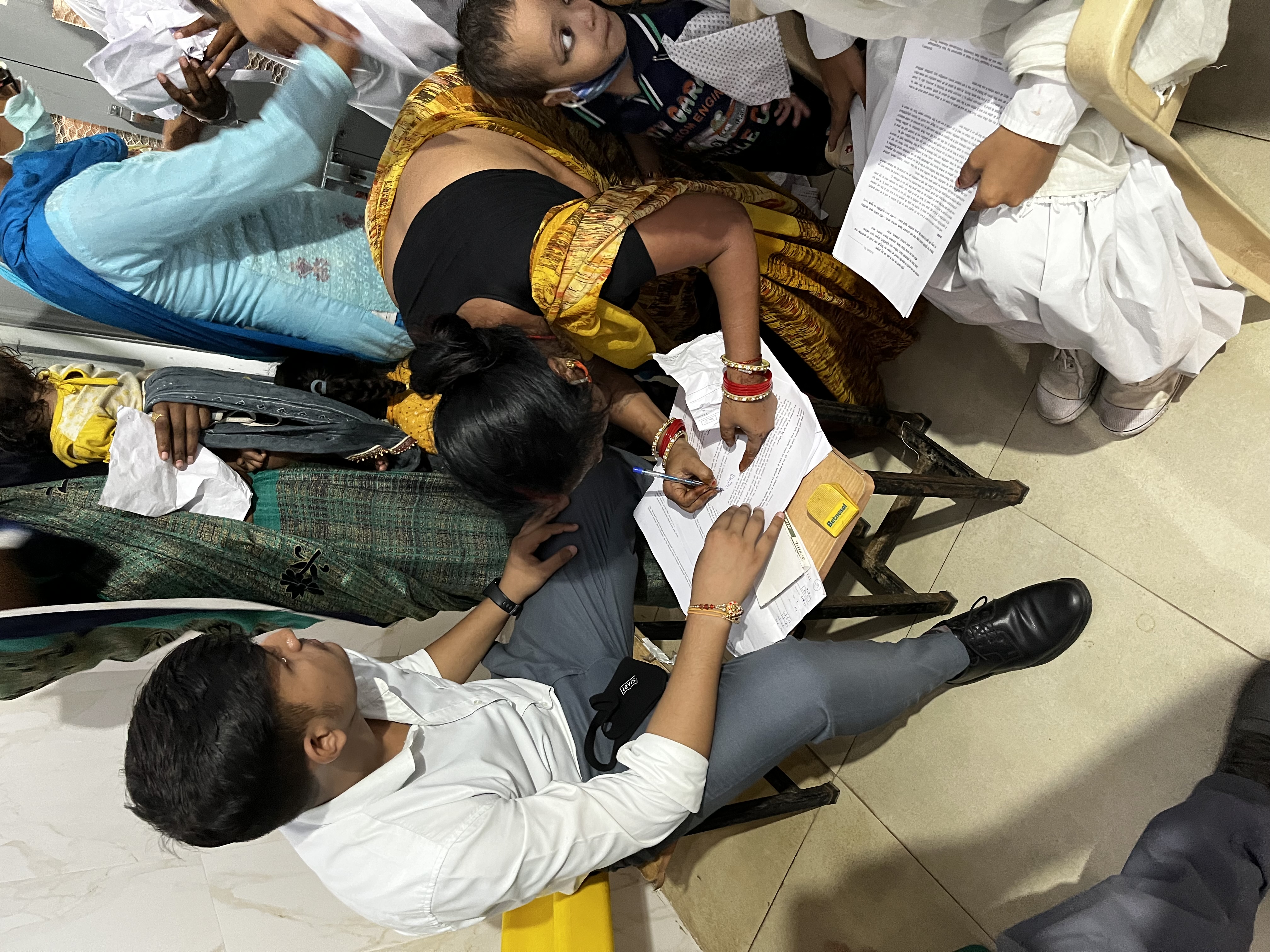}
\caption{Parents signing the consent form after they were explained the objective of the research study, how the data would be collected, and their rights to opt out of the study and have their child palmprint images and biographic information purged from the database.}
\label{fig:consent_form}
\end{center}
\end{figure}
\vspace{-1em}
\section{Database}
\par As mentioned, there is a shortage of palmprint databases of children in the public domain. The authors of this paper organized a data collection camp to collect palmprint data of children between the ages of 9 months to 5 years of age in collaboration with the Dayalbagh Education Institute in Agra, India. Over the course of 5 days starting August 23, 2022, roughly 19,000 palmprint images were collected from 1,027 palms (515 children), out of which demographic labels were collected for 444 subjects. After  cleaning the raw images manually and with the aid of PalmMobile SDK by removing blurry images, accidental captures, failure-to-enroll cases, etc, the cleaned dataset consists of 18,036 images.

\subsection{Data collection protocol}

\par Each subject that agreed to have their data collected were required to sign a consent form approved by the ethics board of the authors' institutional board as well as that of the Saran Ashram Hospital. A volunteer explained the purpose of this project and the data collection protocol itself to the parents or legal guardians of the young subjects (fig. \ref{fig:consent_form}) after which another trained volunteer, capable of communicating in the local language, proceeded to capture the data of the child under the strict supervision of the authors of this study. To collect the data, we use the PalmMobile SDK developed by Armatura. It is a mobile application that is capable of enrolling palms as well as matching them. Figure \ref{fig:collage} shows images captured using this mobile application.


\begin{figure}[t]
\begin{center}
\includegraphics[scale=0.5]{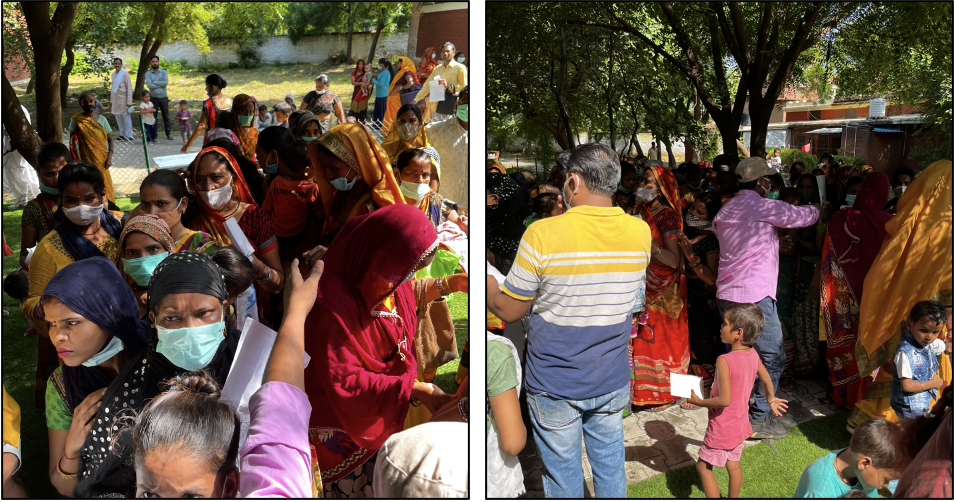}
\caption{Parents with their children lining up outside the data collection site in Dayalbagh, Agra, India.}
\label{fig:lines}
\end{center}
\end{figure}

\begin{figure}[t]
\begin{center}
\includegraphics[scale=0.5]{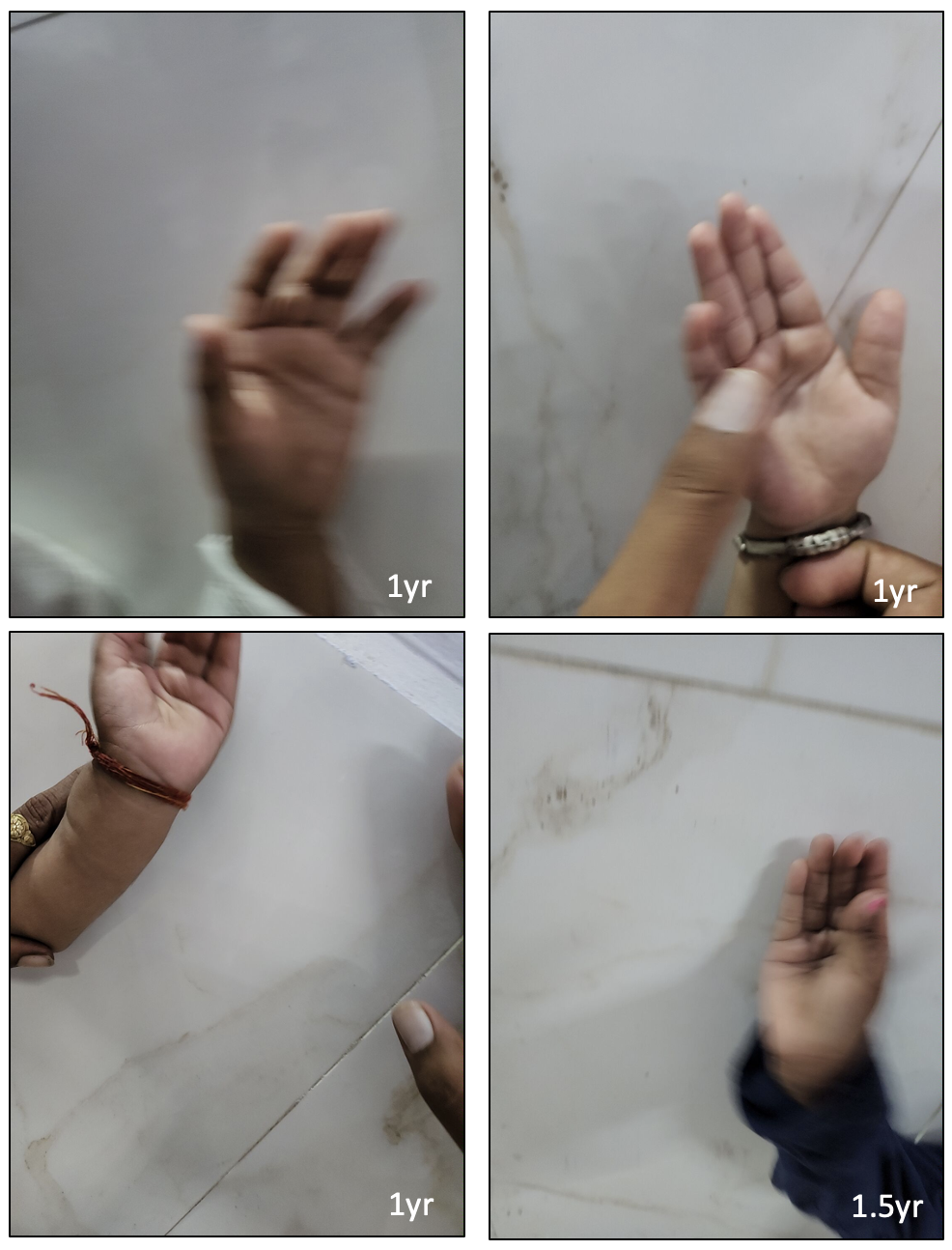}
\caption{Examples of palmprint images that were deleted from our database. These include blurred images, images captured when the child moved their palm unexpectedly or was uncooperative. Out of a total of 19,139 images from 1,027 unique palms, 1,103 images were removed leaving a total of 18,036 images in the cleaned dataset comprising of 1,018 unique palms.}
\label{fig:pruned}
\end{center}
\end{figure}

\par Collecting data at this scale of young children is a challenge since children may be uncooperative while collecting the data or may not have developed the muscles in their hands to fully open their palms. In these cases, we requested the parents of the children to help hold their hand open so that acceptable quality images may be captured of the palms. Additionally, the trained operators were able to communicate with the child better than the authors of the paper which increased the cooperation of the child in many instances. The operator had to constantly communicate with the child to help calm them down so that the palms would be steady for capture (Figure \ref{fig:data_collection}).
\par The authors of this paper had arranged for incentives for the family of each child whose data was collected. A majority of the subjects that participated in the data collection were from underprivileged communities and these incentives provided support for their daily household needs. 

\subsection{Data cleaning}

\par Since there was some level of uncooperation from the young children involved in the data collection process, it is essential to clean up the data set to remove any images that are severely blurred, do not have a palm in them, accidentally captured while the child is being assisted to hold the palm steady etc. Fig \ref{fig:pruned} shows examples of such images that are not included in the cleaned dataset.

\begin{figure}[t]
\begin{center}
\includegraphics[scale=0.55]{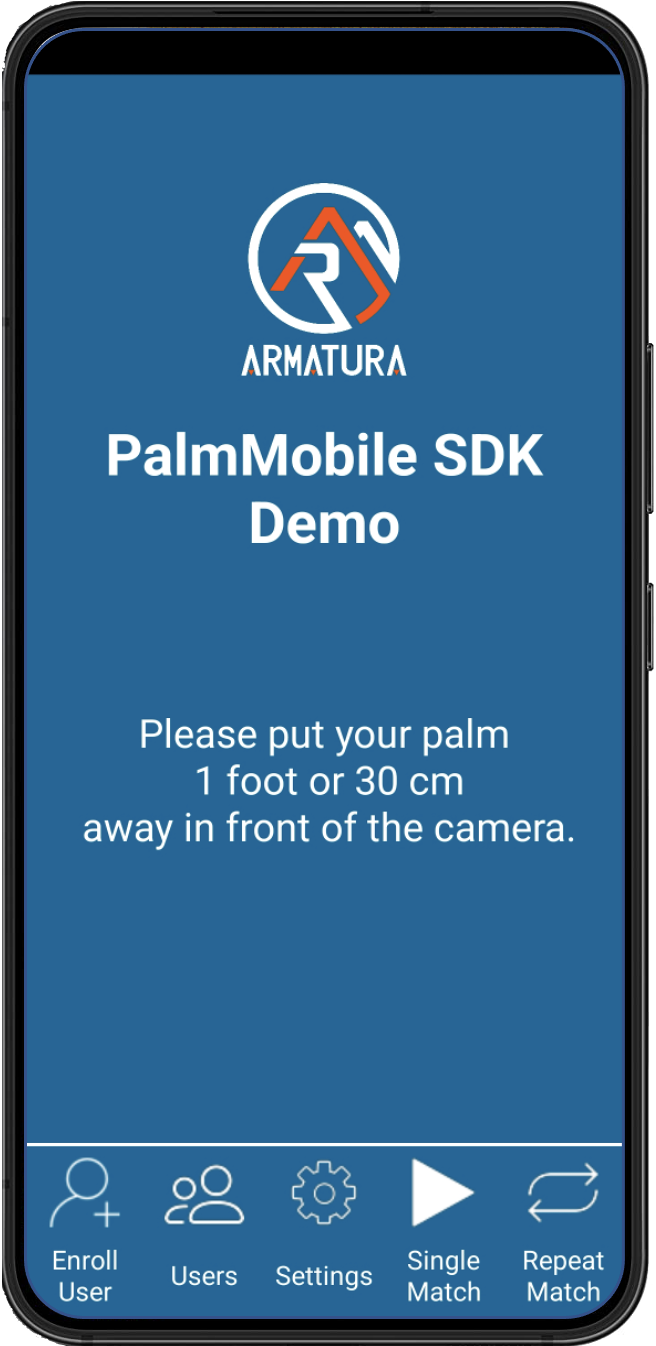}
\caption{User interface of PalmMobile SDK \cite{armatura}.}
\label{fig:demo}
\end{center}
\end{figure}

\section{Experimental Results}
\begin{figure}[t]
\begin{center}
\includegraphics[scale=0.6]{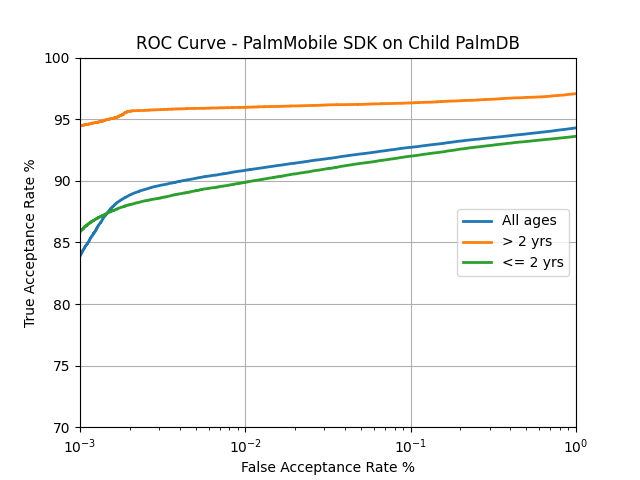}
\caption{Receiver Operating Characteristic Curves partitioned by age group. There are 322 unique palms above the age of 2 years, 566 unique palms below the age of 2 years and 1,018 unique palms overall. For 2 years and under, there are 107,796 genuine comparisons and 93,779,038 impostor comparisons. For ages above 2, there are 29,216 genuine comparisons and 27,054,995 impostor scores. Overall, there are 170,481 genuine comparisons and 171,723,130 impostor scores.}
\label{fig:age_group_roc}
\end{center}
\end{figure}


\par In this report, we evaluated PalmMobile SDK provided by Armatura. This is the same application used for the collection of Child PalmDB. Additionally, we also report results on the Tongji adult palmprint database \cite{zhang2017towards}. This database consists of 12,000 contactless palm images from 600 unique adult palms. Figure \ref{fig:demo} shows the user interface of the PalmMobile SDK mobile app. It has integrated capabilities for user enrolment and palmprint identification, which is extremely convenient for on-site data collection.
\par We evaluate PalmMobile SDK in verification, closed-set identification and open-set identification modes on our Child PalmDB. For each evaluation, we report results on the entire dataset as well as subsets split by age and gender.
\subsection{Verification}
\par
\begin{figure*}[t]
\begin{center}
\includegraphics[scale=0.6]{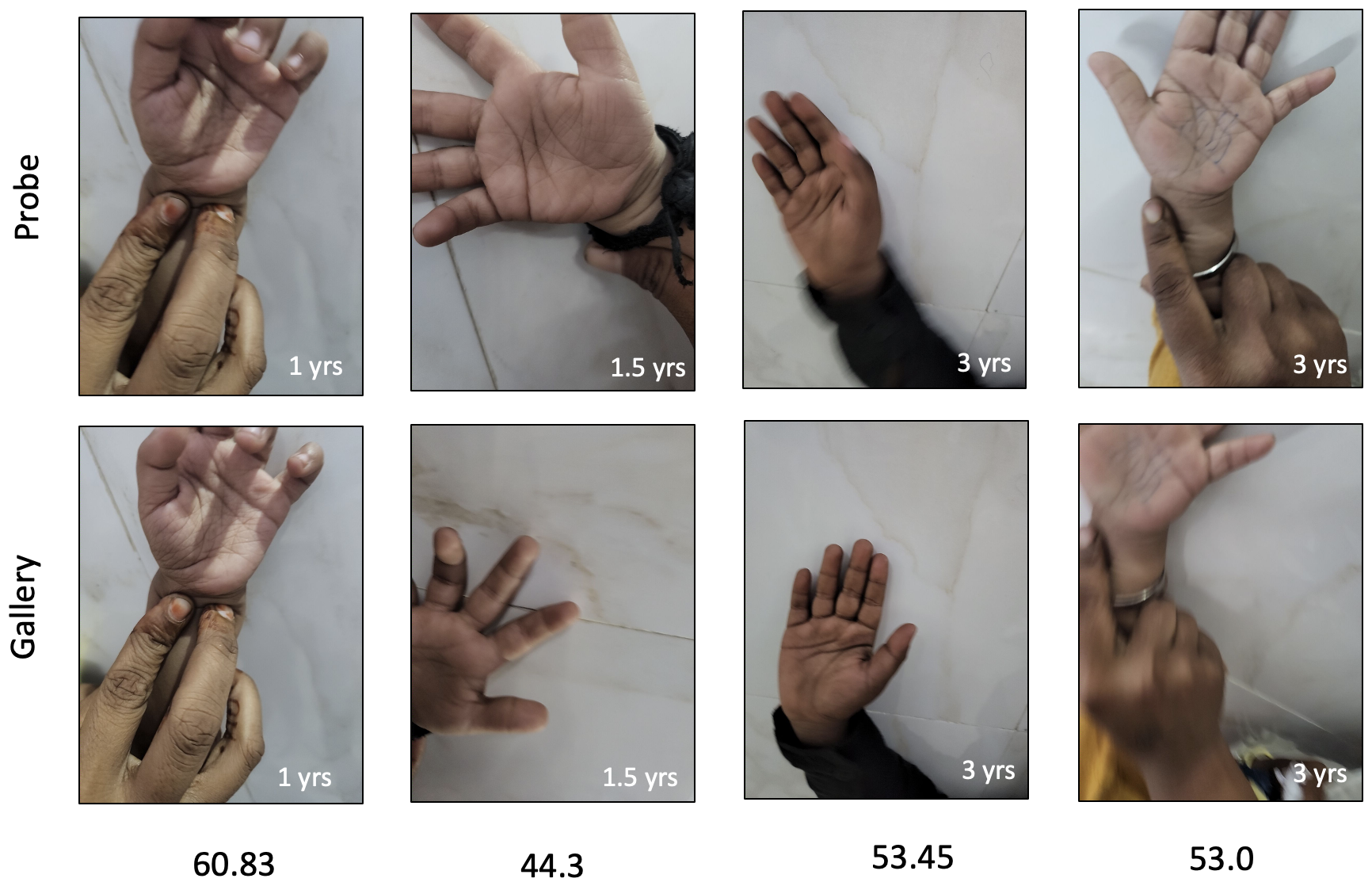}
\caption{Misclassified genuine pairs with the corresponding similarity scores. The similarity score threshold at FAR = 0.01\% is 62.22. The range of scores is [0,100].}
\label{fig:misclassified}
\end{center}
\end{figure*}

 Figure \ref{fig:age_group_roc} shows that the algorithm performs better on  children older than 2 years of age than those below 2 years. This is mainly due to the higher cooperation of older children during the collection of the data and possibly due to a larger number of children below the age of 2 years. Table \ref{tab:eval} summarizes the results for verification. It should be noted that since demographic labels were available for 444 subjects out of 515, the data subsets used in rows 2 and 3 are from 444 subjects. We can see the discrepancy in performance between adult palmprints and child palmprints which is a function of the difficulty in collecting the data for children. Figure \ref{fig:misclassified} shows that the cases in which the matcher fails are those when the two images of the palm are captured at a wide range of angles and when the palm of the child is not stationary and not fully open.

\begin{figure*}[t]
\begin{center}
\includegraphics[scale=0.55]{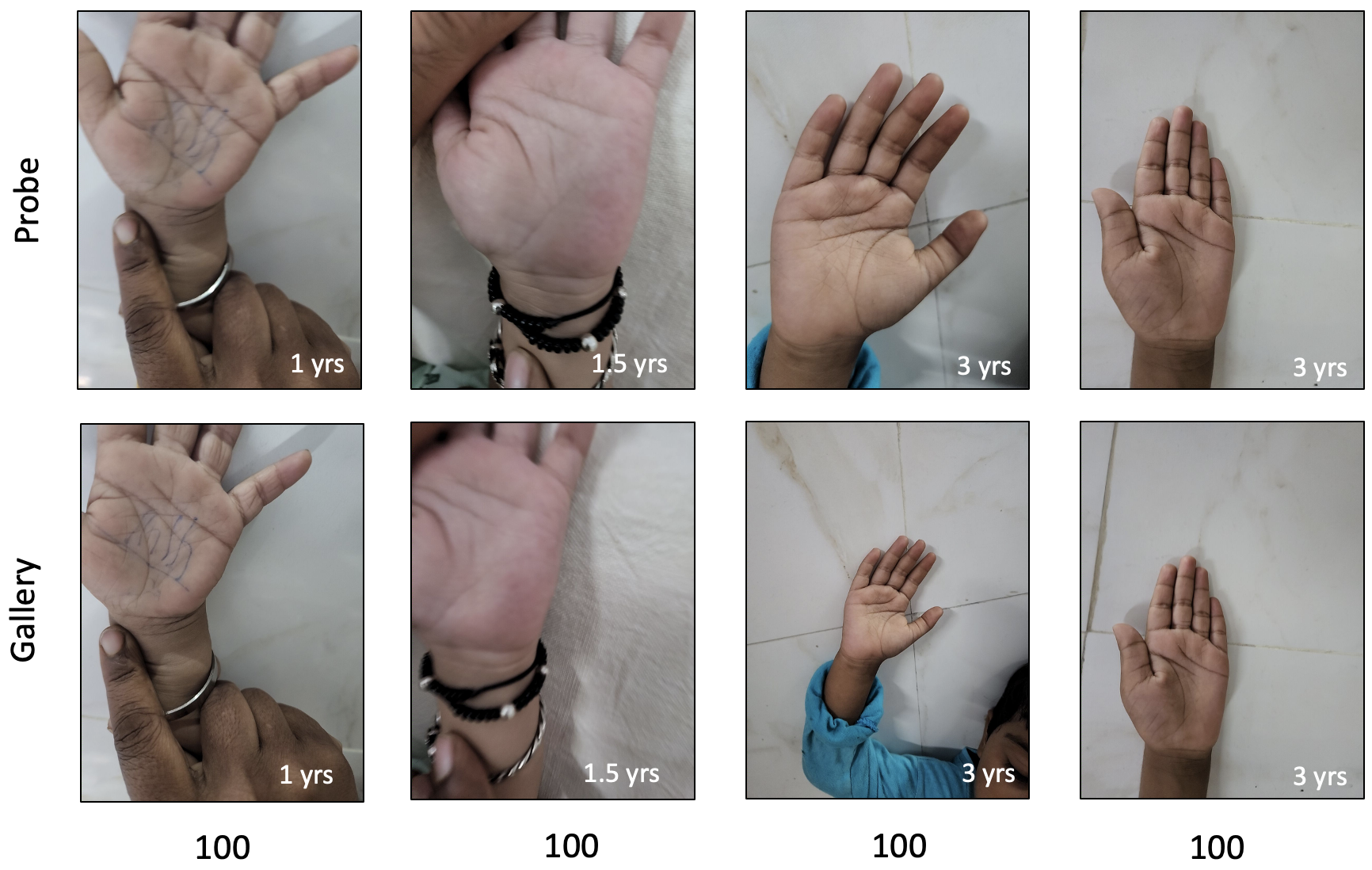}
\caption{Correctly classified genuine pairs with the corresponding similarity scores. The similarity score threshold at FAR = 0.01\% is 62.22. The range of scores is [0,100].}
\label{fig:correctly_classified}
\end{center}
\end{figure*}

\begin{figure*}[t]
\begin{center}
\includegraphics[scale=0.43]{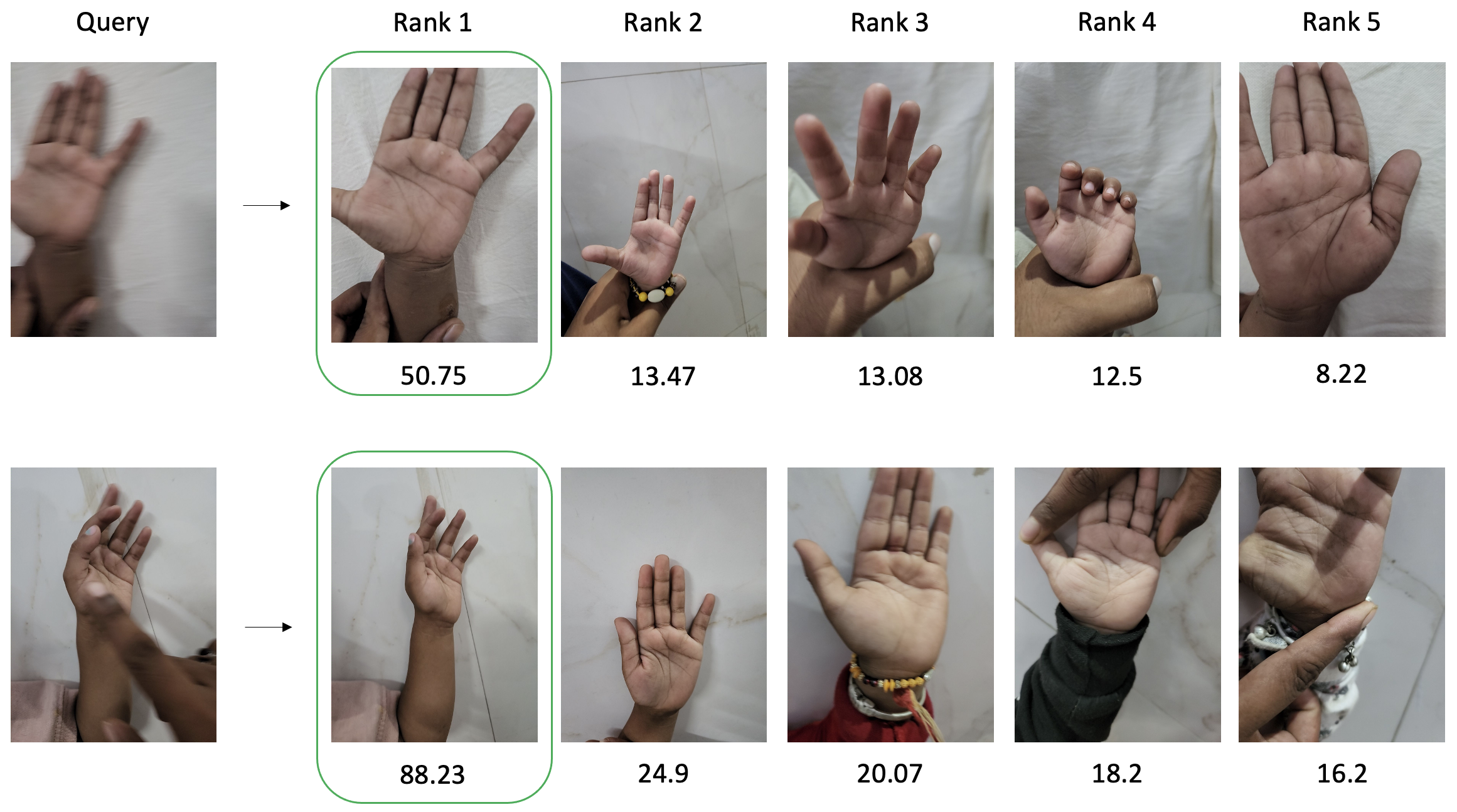}
\caption{Retrievals along with the similarity scores from the closed-set search using one mate per probe in the gallery. Gallery size is 26,636 images.}
\label{fig:retrievals}
\end{center}
\end{figure*}

\subsection{Identification}
\par Since PalmMobile SDK and subsequent palmprint recognition systems will provide the most benefit in an identification setting, it is important to focus on the closed and open-set identification accuracy of the system. Improving the performance in these modes of operation is of primary importance.
\par We report closed set identification results (rank 1) by randomly selecting 200 probe images of unique palms from Child PalmDB. Each of these probes has a single mated identity in the gallery. The gallery comprises of Child PalmDB and Tongji adult palmprints and consists of 26,636 images for Child PalmDB and 26,236 images for Tongji. We employ the same procedure for the Tongji dataset using 200 probes. A similar trend as the verification experiments was observed where the performance was better on older subjects compared to younger subjects. We see better performance in identification compared to verification due to the relatively small gallery size. Table \ref{tab:closed_set} shows that the identification accuracy on adult palmprints is much higher than child palmrpints. We anticipate that this discrepancy will amplify as the gallery size increases. Figure \ref{fig:retrievals} shows the top 5 retrievals for two queries. It is interesting to note that in the case of the first query (top row), the similarity score is below the verification threshold (for FAR = 0.01\%) but the mate was retrieved at rank-1.
\par Additionally, we report the open-set identification results using 100 non-mated and 100 mated probes for Child PalmDB and Tongji adult palmprint dataset. This time, since the probes are non-mated, the gallery size is 26,536 images for Child PalmDB and 26,136 for Tongji. We report the False Positive Identification Rate (FPIR) at a fixed False Negative Identification Rate (FNIR) of 0.3 (30 mated probes out of 200 outside top 5 ranks) in table \ref{tab:open_set}. The performance on adult palmprints was better compared to Child PalmDB. Within Child PalmDB, the performance was better on older subjects compared to younger subjects.

\begin{table}[h]
    \centering
    \caption{TAR(\%) @ FAR = 0.1\% and FAR = 0.01\% using PalmMobile SDK \cite{armatura}}
        \begin{tabular}{|C{0.4\linewidth} | C{0.2\linewidth}|C{0.2\linewidth}|} 
             \hline
             \textbf{Evaluation dataset (\# of subjects)} & \textbf{TAR(\%) \@ FAR = 0.1\%} & \textbf{TAR(\%) \@ FAR = 0.01\%}\\ [1.0ex]
             \noalign{\hrule height 1.2pt}
                Child PalmDB (515)&  92.72 & 90.85\\
             \hline
                Child PalmDB - $\leq$ 2 yrs (283)&  91.99 & 89.88\\ 
             \hline
                Child PalmDB - $>$ 2 yrs (161)&  96.32 & 95.96\\
             \noalign{\hrule height 1.2pt}
             Tongji adult palmprints (300)& 100 & 100 \\
             \hline
        \end{tabular}
    \label{tab:eval}
\end{table}

\begin{table}[h]
    \centering
    \caption{Rank-1 Accuracy for PalmMobile SDK on Child PalmDB and Tongji adult palmprint dataset. Number of mated probes is 200 for Child PalmDB and Tongji database. Gallery size is 26,636 images for Child PalmDB and 26,236 images for Tongji database.}
        \begin{tabular}{|C{0.4\linewidth} | C{0.2\linewidth}|} 
             \hline
             \textbf{Evaluation dataset} & \textbf{Rank-1 Accuracy(\%)}\\ [1.0ex]
             \noalign{\hrule height 1.2pt}
                Child PalmDB & 99.0\\
             \hline
                Child PalmDB - Under 2 years & 98.5\\ 
             \hline
                Child PalmDB - Over 2 years & 100.0\\
             \noalign{\hrule height 1.2pt}
                Tongji adult palmprints & 100.0 \\
            \hline
        \end{tabular}
    \label{tab:closed_set}
\end{table}

\begin{table}[h]
    \centering
    \caption{Open-set identification accuracy for PalmMobile SDK on Child PalmDB and Tongji datasets. Reported as FPIR at FNIR=0.3. Number of mated and non-mated probes is 100 for both Child PalmDB and Tongji database. Gallery size is 26,536 images for Child PalmDB and 26,136 for Tongji.}
        \begin{tabular}{|C{0.4\linewidth} | C{0.2\linewidth}|} 
             \hline
             \textbf{Evaluation dataset} & \textbf{FPIR}\\ [1.0ex]
             \noalign{\hrule height 1.2pt}
                Child PalmDB & 0.01\\
             \hline
                Child PalmDB - Under 2 years & 0.02\\ 
             \hline
                Child PalmDB - Over 2 years & 0.0\\
             \noalign{\hrule height 1.2pt}
                Tongji adult palmprints & 0.0 \\
            \hline
        \end{tabular}
    \label{tab:open_set}
\end{table}

\section{Conclusion and Future Work}
\par A plethora of children around the world continue to suffer and die from vaccine related diseases and malnutrition. A major obstacle standing in the way of delivering the vaccinations and nutrition needed to the children most in need is the means to quickly and accurately identify or authenticate a child at the point of care. To address this challenge, we evaluate a low-cost contactless child Palm-ID, an end-to-end child palmprint recognition system for the age group 1-5 years.  
It is our hope that this report will motivate a strong push for biometric recognition systems which can be used to alleviate child suffering around the world. In doing so, we believe that this work will make a major dent in Goal \#3 of the United Nations Sustainable Development Goals, namely, “Ensuring healthy lives and promoting well-being for all, at all ages.” 

\section{Acknowledgements}
\par We thank Armatura for providing us with the mobile application that was used to collect the data as well as the PalmMobile SDK that was used to perform the evaluations on the Child PalmDB database.
\ifCLASSOPTIONcaptionsoff
  \newpage
\fi

\bibliography{cite}

\begin{thebibliography}{10}

\bibitem{uidai}
UIDAI, ``Aadhaar enrolment - {U}nique {I}dentification {A}uthority of {I}ndia:
  {G}overnment of {I}ndia.''
\newblock
  \href{https://uidai.gov.in/en/my-aadhaar/about-your-aadhaar/aadhaar-enrolment.html}{https://uidai.gov.in/en/my-aadhaar/about-your-aadhaar/aadhaar-enrolment.html}.

\bibitem{jainbiometrics2011}
A.~K. Jain, A.~A. Ross, and K.~Nandakumar, {\em Introduction to Biometrics}.
\newblock Springer Publishing Company, Incorporated, 2011.

\bibitem{gavi}
``Gavi {B}oard responds to an uncertain world: Fragile and conflict settings,
  future pandemics and the ongoing fight against {COVID}-19.''
\newblock
  \href{https://www.gavi.org/news/media-room/gavi-board-responds-uncertain-world-fragile-conflict-settings-future-pandemics}{https://www.gavi.org/news/media-room/gavi-board-responds-uncertain-world-fragile-conflict-settings-future-pandemics}.

\bibitem{programme_2018}
{World Food Programme}, ``5-year old {S}imon having his fingerprints and iris
  scanned with {SCOPE} technology during a {WFP} registration in {South Sudan}.
  {B}iometric verification is helping us make sure that the right assistance
  reaches the right people.,'' Jul 2018.
\newblock
  \href{https://twitter.com/wfp/status/1017663925368901632}{https://twitter.com/wfp/status/1017663925368901632}.

\bibitem{engelsma2021infant}
J.~J. Engelsma, D.~Deb, K.~Cao, A.~Bhatnagar, P.~S. Sudhish, and A.~K. Jain,
  ``Infant-id: Fingerprints for global good,'' {\em IEEE Transactions on
  Pattern Analysis and Machine Intelligence}, vol.~44, no.~7, pp.~3543--3559,
  2021.

\bibitem{lemes2011biometric}
R.~P. Lemes, O.~R. Bellon, L.~Silva, and A.~K. Jain, ``Biometric recognition of
  newborns: Identification using palmprints,'' in {\em 2011 International Joint
  Conference on Biometrics (IJCB)}, pp.~1--6, IEEE, 2011.

\bibitem{press_info}
``Aadhaar 2.0- ushering the next era of digital identity and smart governance
  workshop from 23rd to 25th november, 2021.''
\newblock
  \href{https://pib.gov.in/PressReleaseIframePage.aspx?PRID=1775073}{https://pib.gov.in/PressReleaseIframePage.aspx?PRID=1775073}.

\bibitem{jain2016fingerprint}
A.~K. Jain, S.~S. Arora, K.~Cao, L.~Best-Rowden, and A.~Bhatnagar,
  ``Fingerprint recognition of young children,'' {\em IEEE Transactions on
  Information Forensics and Security}, vol.~12, no.~7, pp.~1501--1514, 2016.

\bibitem{liu2017infant}
E.~Liu, ``Infant footprint recognition,'' in {\em Proceedings of the IEEE
  International Conference on Computer Vision}, pp.~1653--1660, 2017.

\bibitem{ramachandra2018verifying}
R.~Ramachandra, K.~B. Raja, S.~Venkatesh, S.~Hegde, S.~D. Dandappanavar, and
  C.~Busch, ``Verifying the newborns without infection risks using contactless
  palmprints,'' in {\em 2018 International Conference on Biometrics (ICB)},
  pp.~209--216, IEEE, 2018.

\bibitem{yambay2019feasibility}
D.~Yambay, M.~Johnson, K.~Bahmani, and S.~Schuckers, ``A feasibility study on
  utilizing toe prints for biometric verification of children,'' in {\em 2019
  International Conference on Biometrics (ICB)}, pp.~1--7, IEEE, 2019.

\bibitem{saggese2019biometric}
S.~Saggese, Y.~Zhao, T.~Kalisky, C.~Avery, D.~Forster, L.~E. Duarte-Vera, L.~A.
  Almada-Salazar, D.~Perales-Gonzalez, A.~Hubenko, M.~Kleeman, {\em et~al.},
  ``Biometric recognition of newborns and infants by non-contact
  fingerprinting: lessons learned,'' {\em Gates Open Research}, vol.~3, 2019.

\bibitem{arcieri_2021}
K.~Arcieri, ``Amazon {E}xpands {P}alm-{R}eading {P}ayment {S}ystem as
  {P}andemic {D}rives {D}own {C}ash {U}se,'' Feb 2021.
\newblock
  \href{https://www.spglobal.com/marketintelligence/en/news-insights/latest-news-headlines/amazon-expands-palm-reading-payment-system-as-pandemic-drives-down-cash-use-62545416}{https://www.spglobal.com/marketintelligence/en/news-insights/latest-news-headlines/amazon-expands-palm-reading-payment-system-as-pandemic-drives-down-cash-use-62545416}.

\bibitem{publisher_2022}
J.~M. Germain, ``New {C}ontactless {B}iometric {S}ystem {U}ses {H}ands as
  {S}ecure {P}asswords,'' Feb 2022.
\newblock
  \href{https://www.crmbuyer.com/story/new-contactless-biometric-system-uses-hands-as-secure-passwords-87404.html}{https://www.crmbuyer.com/story/new-contactless-biometric-system-uses-hands-as-secure-passwords-87404.html}.

\bibitem{matrix_access_control_2022}
{Matrix Access Control}, ``Facts {A}bout {P}alm {V}ein {R}ecognition {S}ystem -
  {A}ll {Y}ou {N}eed to {K}now,'' Aug 2022.
\newblock
  \href{https://www.matrixaccesscontrol.com/blog/facts-about-palm-vein-recognition-system-all-you-need-to-know/}{https://www.matrixaccesscontrol.com/blog/facts-about-palm-vein-recognition-system-all-you-need-to-know/}.

\bibitem{dian2016contactless}
L.~Dian and S.~Dongmei, ``Contactless palmprint recognition based on
  convolutional neural network,'' in {\em 2016 IEEE 13th International
  Conference on Signal Processing (ICSP)}, pp.~1363--1367, IEEE, 2016.

\bibitem{zhang2017towards}
L.~Zhang, L.~Li, A.~Yang, Y.~Shen, and M.~Yang, ``Towards contactless palmprint
  recognition: A novel device, a new benchmark, and a collaborative
  representation based identification approach,'' {\em Pattern Recognition},
  vol.~69, pp.~199--212, 2017.

\bibitem{liu2020contactless}
Y.~Liu and A.~Kumar, ``Contactless palmprint identification using deeply
  learned residual features,'' {\em IEEE Transactions on Biometrics, Behavior,
  and Identity Science}, vol.~2, no.~2, pp.~172--181, 2020.

\bibitem{morales2011towards}
A.~Morales, M.~A. Ferrer, and A.~Kumar, ``Towards contactless palmprint
  authentication,'' {\em IET Computer Vision}, vol.~5, no.~6, pp.~407--416,
  2011.

\bibitem{wu2014sift}
X.~Wu, Q.~Zhao, and W.~Bu, ``A {SIFT}-based contactless palmprint verification
  approach using iterative {RANSAC} and local palmprint descriptors,'' {\em
  Pattern Recognition}, vol.~47, no.~10, pp.~3314--3326, 2014.

\bibitem{leng2017dual}
L.~Leng, M.~Li, C.~Kim, and X.~Bi, ``Dual-source discrimination power analysis
  for multi-instance contactless palmprint recognition,'' {\em Multimedia Tools
  and Applications}, vol.~76, no.~1, pp.~333--354, 2017.

\bibitem{mascellino_2021}
A.~Mascellino, ``Fujitsu unveils multimodal biometric authentication technology
  for contactless retail: Biometric update,'' Jan 2021.

\bibitem{rajaram2022palmnet}
K.~Rajaram, A.~Devi, and S.~Selvakumar, ``Palmnet: A cnn transfer learning
  approach for recognition of young children using contactless palmprints,'' in
  {\em Machine Learning and Autonomous Systems}, pp.~609--622, Springer, 2022.

\bibitem{armatura}
``Products powered by {A}rmatura.''
\newblock
  \href{https://armatura.us/products/#palmrec}{https://armatura.us/products/\#
  palmrec}.

\end{thebibliography}
\bibliographystyle{ieeetr}



\end{document}